\pdfoutput=1

\documentclass[11pt]{article}

\usepackage{acl}

\usepackage{times}
\usepackage{latexsym}

\usepackage[T1]{fontenc}

\usepackage[utf8]{inputenc}

\usepackage{microtype}

\usepackage{inconsolata}

\usepackage{soul}
\usepackage{amsmath}
\usepackage{amsfonts}
\usepackage{multirow, multicol}
\usepackage{siunitx}
\usepackage{graphicx}

\usepackage{tabularx, makecell}
\usepackage{xspace}
\usepackage{booktabs,arydshln}

\usepackage{algorithm}
\usepackage[noend]{algpseudocode}

\makeatletter
\def\adl@drawiv#1#2#3{%
        \hskip.5\tabcolsep
        \xleaders#3{#2.5\@tempdimb #1{1}#2.5\@tempdimb}%
                #2\z@ plus1fil minus1fil\relax
        \hskip.5\tabcolsep}
\newcommand{\cdashlinelr}[1]{%
  \noalign{\vskip\aboverulesep
           \global\let\@dashdrawstore\adl@draw
           \global\let\adl@draw\adl@drawiv}
  \cdashline{#1}
  \noalign{\global\let\adl@draw\@dashdrawstore
           \vskip\belowrulesep}}
\makeatother

\newcommand{\deberta}{\textsc{DeBERTa}\xspace}
\newcommand{\bart}{BART\xspace}
\newcommand{\DKNN}{\textsc{DkNN}\xspace}
\newcommand{\toxigen}{\textsc{ToxiGen}\xspace}
\newcommand{\esnli}{\textsc{E-SNLI}\xspace}

\newcommand{\increase}[1]{\textcolor{NavyBlue}{#1}}
\newcommand{\decrease}[1]{\textcolor{Maroon}{#1}}
\newcommand{\premise}{\underline{Premise:}\xspace}
\newcommand{\hypothesis}{\underline{Hypothesis:}\xspace}
\newcommand{\human}{\underline{Human explanation:}\xspace}
\newcommand{\knn}{$k$NN\xspace}
\newcommand{\lmeans}{$L$-means\xspace}

\usepackage[textsize=scriptsize, disable]{todonotes}
\newcommand\sam[1]{\todo[color=blue!20]{{\bf Sam}: #1}}

\newcommand\ml[1]{\todo[color=red!20]{{\bf Matt}: #1}}

\soulregister{\citep}{1}
\soulregister{\citet}{1}
\soulregister{\cite}{1}

\title{Wrapper Boxes: Faithful Attribution of Model Predictions to Training Data}

\author{Yiheng Su \and Junyi Jessy Li \and Matthew Lease \\
        The University of Texas at Austin \\
        \texttt{\{sam.su, jessy, ml\}@utexas.edu}}

\begin{document}
\maketitle
\begin{abstract}
Can we preserve the accuracy of neural models while also providing {\em faithful} explanations of model decisions to training data? We propose a ``wrapper box'' pipeline: training a neural model as usual and then using its learned feature representation in classic, interpretable models to perform prediction. 
Across seven language models of varying sizes, including four large language models (LLMs), 
two datasets at different scales, three classic models, and four evaluation metrics, we first show that predictive performance of wrapper classic models is largely comparable to the original neural models. 

Because classic models are transparent, each model decision is determined by a known set of training examples that
can be directly shown to users. Our pipeline thus preserves the predictive performance of neural language models while faithfully attributing classic model decisions to training data. 
Among other use cases, such attribution enables model decisions to be contested based on responsible training instances. Compared to prior work, our approach achieves higher coverage and correctness in identifying which training data to remove to change a model decision. 
To reproduce findings, our source code is online at: \url{https://github.com/SamSoup/WrapperBox}. 
\end{abstract}

\begin{figure*}[t]
    \centering
    \includegraphics[scale=0.6]{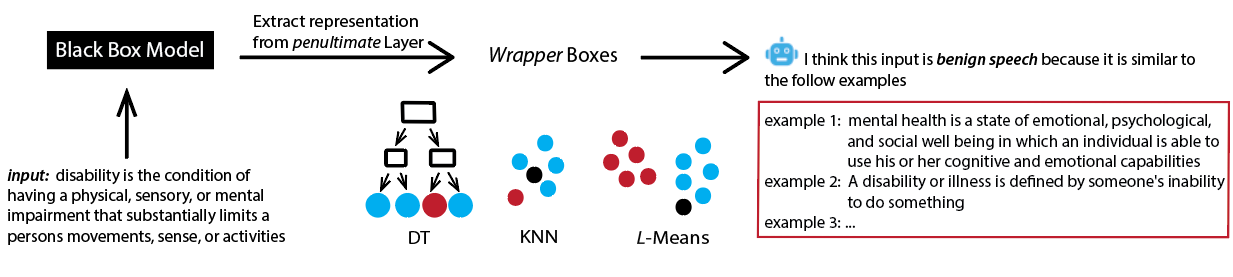}
    \vspace{-1.5em}
    \caption[WrapperBox]{Three wrapper boxes 
     illustrated for toxic language detection  \citep{hartvigsen2022toxigen}. 
Red and blue dots denote harmful vs.\ benign speech. Smaller dots represent examples, while larger dots represent clusters (e.g., DT leaf nodes). A neural model's penultimate layer provides the feature representation for 
the white wrapper boxes. Our results show that classic models can achieve comparable performance to the underlying neural models while 
also providing intuitive, example-based explanations (described in Section~\ref{sec:whiteboxes}).}
    \label{fig:WrapperBox}
    \vspace{-1em}
\end{figure*}

\section{Introduction}
\vspace{-0.5em}
Opaque predictive models are challenging to trust and reason about, prompting calls for greater transparency and interpretability in automated decisions \citep{Langer2021-ek, Shin2021-rj}. In critical sectors like law, health, and finance, interpretability may be essential to prevent catastrophic failures \citep{Ahmad2018-gk, rudin2019stop, Bhatt2020-rc}. 
Furthermore, interpretability may be required for regulatory compliance \citep{Kaminski2019-cf}. 
However, popular pre-trained language models \citep{devlin2018bert, lewis2019bart, floridi2020gpt, chung2022scaling} are inscrutable, making it difficult to explain model decisions \citep{Adadi2018-wx, Barredo_Arrieta2020-ym}. 

%


In contrast,
classic ``white box'' methods such as $k$-nearest neighbor (\knn) and decision tree (DT) are inherently interpretable \cite{rudin2019stop}: each model decision is determined by a known set of training examples that can be
directly shown to users. Nevertheless, classic models tend to underperform today's neural models. 
%

Recent work has pursued ways to blend the interpretability of classic models with the predictive performance of today's neural models \cite{wang2017k, wangTree2018, papernot2018deep, wallace-etal-2018-interpreting, rajani2019explain, rajagopal2021selfexplain}. 
However, prior models face limitations in efficiency and scalability, requiring training from scratch or expensive computation and storage. 
%

In addition, research on interpretable NLP has largely focused on feature-style explanations, with far less work on {\em example-based explanations} \citep{KeaneCaseBased}. Because people naturally reason by analogy  \cite{Sormo2005-gx, Schank2014-bf, Kolodner2014-ch}, explaining predictions to specific training data is intuitively appealing. Example-based explanations also connect to work on case-based reasoning \cite{aamodt1994case} by relating new problem instances to similar past ones, a problem-solving strategy people naturally use in decision-making \citep{newel1972human}. \citet{rudin2019grandchallenge} thus argues for developing modern case-based methods as one of the grand challenges in interpretable machine learning. 

%
%
%


In this work, we synthesize existing black-box and white-box methods toward building (training data) attributable-by-design models. Specifically, we introduce the {\em wrapper box} pipeline to combine the accuracy of modern neural models with the faithful, example-based explanations of classic models. Our approach effectively ``wraps'' a given neural model with one or more transparent classic models to maintain neural performance while improving interpretability. Building on the tradition of fitting fully connected layers on neural representations \cite{AlexNet2012, simonyan2015deepVGG, he2016deep, devlin2018bert}, we fit white classic models on extracted neural representations for inference. The reasoning process of a classic wrapper model can then be faithfully explained by showing the specific training examples that led to each prediction ({\bf Figure~\ref{fig:WrapperBox}}).

Note that wrapper boxes do {\em not} attempt to approximate the underlying neural model, i.e., wrapper box vs.\ neural model predictions can differ. Our claim is that wrapper box predictions can be faithfully explained and that the predictive performance of wrapper boxes is largely comparable to the underlying neural models (see below).  


In contrast with prior techniques for generating example-based explanations ({\bf Table~\ref{tab:comp}}), wrapper box explanations are fully faithful to how the actual predictions are made and do not require additional neural training or high run-time stipulations. The wrapper box concept is also quite general, as we show across three pre-trained language models ({\bart-large}, {\deberta-large}, and Flan-T5-large) and three classic models: \knn, DT, and $k$-means. 






%


Our first evaluation assesses the predictive performance of wrapper boxes on two classification tasks with varying data scales: toxic speech detection ({\textsc{\toxigen}}) and natural language inference ({\textsc{\esnli}}). 
We show that wrapper box predictive performance is largely comparable to neural baselines. While some statistically significant differences are observed ($12$\%-- $27$\% across our datasets), this also includes cases in which the wrapper box actually performs significantly better than the base neural model. In the few cases where performance is worse, the value of interpretability may still justify use, bolstered by the vast majority of cases where no statistically significant difference is observed. 

We also evaluate the effectiveness of wrapper boxes using representations from modern large language models (LLMs). Namely, we experiment with Llama 2-7B Instruct \citep{touvron2023llama}, Llama 3-8B-Instruct \citep{dubey2024llama}, Mistral-7B Instruct \cite{Jiang2023Mistral7B}, and Gemma-7B Instruct \cite{Mesnard2024Gemma}. Results show that wrapper boxes using zero-shot LLM representations strongly outperform baseline LLM performance across both tasks. 

Next, we demonstrate the usefulness of example-based explanations from wrapper boxes
to attribute model decisions to specific training data. 
Such attribution supports intuitive model explanations for end-users \citep{Schank2014-bf} and enabling {\em data-centric} approaches for model developers, such as data cleaning \citep{Zylberajch2021-ym}.

\begin{table*}[t]
    \centering
    \small
    \begin{tabular}{p{4cm}p{10.5cm}}
    \toprule
    Prototypes & Full network training necessary. \\
    \cite{das2022prototex} & Fully or partially faithful by retrieving examples closest to learned prototypes. \\ \midrule

    Concepts & Full network training necessary. \\
    \cite{rajagopal2021selfexplain} & Partially faithful with learned concepts in interpretability layers.
    \\ \midrule

    Influence functions & No training but high runtime $\mathcal{O}(np^2 +p^3)$ (n: dataset size, p: model parameters). \\
    \cite{koh2017understanding} & Not faithful (post-hoc estimate only) but agnostic to the underlying architecture. \\ \midrule

    DkNN & No training but high runtime and storage requirements \\
    \cite{papernot2018deep} & Fully or partially faithful depending on nearest neighbors shown. 
    \\ \midrule
    
    {\bf Wrapper Boxes} & No neural model retraining; classic wrapper box models are trained as appropriate. \\
    (this work) & Fully or partially faithful depending on 
    examples shown, agnostic to representations.
    \\

    \bottomrule
   \end{tabular}
   \caption{A comparison of this work among closest prior works in example-based explanations. Note that loss in fidelity for wrapper boxes can only occur by conscious decision, e.g., if one chooses to show fewer training examples than were used during inference to simplify explanations for end-users further (see Case II from Table~\ref{tab:Strategies}).}
\label{tab:comp}
\vspace{-1em}
\end{table*}

Finally, we evaluate another use case: enabling model decisions to be contested based on the training data responsible for those decisions. Specifically, we consider identifying which training data needs to be removed to change a model decision \cite{Yang2023-st}. This task offers a form of {\em algorithmic recourse} \citep{Karimi2022-cz}, which emphasizes providing actionable explanations to users unfavorably treated by automated systems. Users are provided a foundation for contesting model decisions by attributing model decisions to specific training data. Compared to \hbox{\citet{Yang2023-st}}, we show higher coverage and correctness in identifying which training data to remove while also generalizing beyond simple linear models and scaling to more modern neural networks.


\section{Related Work}
\label{sec:RW}

\paragraph{Explainable Models}

Most work on interpretability focuses on post hoc methods that explain a pre-trained model retroactively \citep{madsen2022post}. This includes input attribution \citep{ribeiro2016should, wangTree2018, mosca-etal-2022-shap, NielsenGradient2022} and attention-based \citep{serrano-smith-2019-attention, sun-lu-2020-understanding} methods. Others seek to design inherently interpretable \citep{rudin2019stop, sudjianto2021designing} models instead, such as prototype networks \citep{das2022prototex, wen2023few}. 
%
Post hoc methods are more versatile and readily applicable but can lead to unfaithful or misleading explanations \citep{basu2021influence, zhang2021sample}. On the other hand, inherently interpretable methods 
offer faithful explanations but may sacrifice performance \citep{du2019techniques}.

\paragraph{Example-based Explanations}

Both input attribution and example-based explanations seek to explain model predictions in relation to observable data (i.e., inputs and training examples) rather than latent representations. This allows feature representations to be optimized for predictive performance without complicating explanations for end-users. 

Unlike input attribution methods, example-based explanations \citep{KeaneCaseBased} aim to identify similar training inputs as analogical justification for model predictions. 
Early work \citep{caruana1999case} proposed treating the activation patterns of hidden nodes in a multi-layer perceptron as features for 1-nearest neighbor and decision trees. Most prior work offers post hoc case-based reasoning via influence functions that show the training points most critical to a specific prediction as explanations \citep{koh2017understanding, han2020explaining, wallace2020interpreting, pruthi2020estimating}. \citet{rajagopal2021selfexplain} offer an inherently interpretable model, although derived concepts (non-terminal phrases) for explanation are at best partially faithful. {\bf Table~\ref{tab:comp}} compares our work with the most similar example-based approaches in prior work. 
\ml{all recent we know of?}

Prior work has consistently validated the significance, utility, and effectiveness of example-based explanations \cite{aamodt1994case, Sormo2005-gx, richter2016case}. 
%
Benefits for users include increased model understanding (simulatability), complementary performance, and trust. \cite{Yeh2018-pt, papernot2018deep, Cai2019-kq, hase2020evaluating, han2020explaining, rajagopal2021selfexplain, das2022prototex, suresh2022intuitively, Chen2023-ft}.
For developers, tying inference to specific training examples can uncover artifacts \citep{lertvittayakumjorn2021explanation}, errors \citep{koh2017understanding},
%
%
and gaps \citep{Khanna2019-ho} in training data, which can be addressed by label cleaning \cite{teso2021interactive},
data augmentation \cite{feng2021survey}, and other {\em data-centric} techniques \cite{Anik2021-xw}.

These studies show that example-based explanations are especially effective in the vision and text domains, given the intuitive nature of images and words \cite{CarvalhoInterpretability2019}. Furthermore, in health and law, where decisions rely on historical precedents, case-based reasoning can assist users in developing intuitions for a model's inference procedure \citep{AYOUB2021102569, zhou2021evaluating}. Of course, if training data is private, then example-based explanations are not possible \citep{dodge2022position}. Section~\ref{sec:DataTypes} further examines the suitability of example-based explanations.

While some studies have reported other forms of explanation being preferred over case-based explanations \citep{Binns2018UserCBE, Dodge2019userCBE, Wang2021-ui}, none of the case-based systems evaluated provided faithful explanations. 

\paragraph{Deep \knn}

We build on \citet{papernot2018deep}'s \DKNN, which has been applied to text classification~\citep{wang2017k, wallace-etal-2018-interpreting,rajani2020explaining}.
%
%
Our work generalizes \DKNN both conceptually and empirically to a broader suite of wrapper box models: 
decision trees (DTs) and clustering-based classification alongside \knn. This differs from prior approaches, which rely on traditionally learned linear components to forecast decisions \cite{koh2017understanding, rajagopal2021selfexplain, das2022prototex}. 
Our framework is arguably the easiest to understand, implement, and reproduce. We do not modify the original model 
nor require additional computation beyond fitting white boxes and additional pass over training data to extract representations (which may be done offline). We thus avoid expensive operations required by prior work, such as approximating inverted Hessian gradients \citep{koh2017understanding} or training a network from scratch \citep{rajagopal2021selfexplain, das2022prototex}. 
Unlike prior work, the wrapper box framework is designed to be dataset, model, and task-agnostic. 

\paragraph{Model Auditing/Algorithmic Recourse}

Model auditing and algorithmic recourse are commonly cited goals for fair and accountable AI systems and, thus, are closely related to explainability \citep{Deck2024-vv}. Model auditing \citep{Bandy2021-jv, Brown2021-jw, Yang2023-st} involves systematically examining a model's behavior to identify problematic behaviors, potential biases, and errors in the training data. A natural next step is algorithmic recourse \citep{Karimi2022-cz}, which emphasizes providing actionable explanations and recommendations to users unfavorably treated by automated systems. By faithfully attributing model decisions to specific training data, 
wrapper boxes provide an avenue 
for contesting unjust decisions to support algorithmic recourse. 



\begin{table*}[t]
    \centering
    \small
    \begin{tabular}{c| c c c c c}
     Case & Influential Examples & Explanation Examples & Performance & Faithfulness & Simplicity \\
    \toprule
    I & All Relevant & All Relevant & $\checkmark$ & $\checkmark$ & $\circ$ \\
    II & All Relevant & Subset & $\checkmark$ & $\circ$ & $\checkmark$ \\
    III & Subset & Subset & $\circ$ & $\checkmark$ & $\checkmark$ \\
    \end{tabular}
    \caption{Case-based models permit tradeoffs between key outcome variables -- predictive performance, explanation faithfulness (or fidelity), and explanation simplicity -- based on which (influential) training examples are used in making a prediction vs.\ to explain that prediction. Note that for any given input, different models will naturally vary in which training examples are influential in performing inference for that input.
}
   \label{tab:Strategies}
   \vspace{-1em}
\end{table*}

\section{Example-based Explanation Tradeoffs}
\label{sec:explain}
\vspace{-0.5em}



We focus on interpretable predictive models that tie inference directly to specific training examples,  
%
%
enabling each prediction to be faithfully explained via those same training examples that determined the model's prediction. Appendix~\ref{sec:perception} further discusses user perceptions of machine-retrieved examples. \sam{appendix mention here}

To better elucidate the design space for working with such models, this section illustrates possible tradeoffs between three key variables of interest: predictive performance, explanation \textit{faithfulness}, and explanation \textit{simplicity}. Following \citet{jacovi2020towards}, we conceptually define faithfulness as how accurately presented explanations reflect the actual reasoning process of the inference model. Concretely, we evaluate the faithfulness of example-based explanations by \textit{completeness} \citep{gu-etal-2023-iaeval}, where derived examples are faithful to the extent that all instances that support the test prediction are selected. The simplicity of example-based explanations can be intuitively quantified as the number of presented instances \citep{Nguyen2020-ib}. 

Section \ref{sec:whiteboxes} discusses how such tradeoffs can be operationalized in practice for our specific wrapper box models. 

\subsection{Conceptual Tradeoffs}
\label{sec:conceptual}
\vspace{-0.5em}


Given an input, assume the prediction model consults $n$ training examples to make a prediction. Furthermore, assume that $m <= n$ of these training examples are shown to explain the prediction. When $m=n$, this explanation is fully faithful to the actual prediction. However, if $n$ is very large, showing all $m=n$ of these training examples to explain the prediction may induce {\em cognitive overload}, often also referred to as {\em information overload} \cite{Marois2005-oq, Abdul2020-bc}. 


To simplify the explanation, one could reduce it to a smaller subset of $m < n$ of the training examples used in prediction. However, this would compromise explanation fidelity. Alternatively, the number of training examples $n$ used in prediction could be reduced. With a smaller $n$, all $m=n$ examples could be shown, boosting explanation simplicity while preserving fidelity, but possibly at the cost of reduced performance.

{\bf Table~\ref{tab:Strategies}} further illustrates the range of possible tradeoffs by presenting three scenarios, Cases I-III. 


\underline{Case I} attains high predictive performance and explanation fidelity, but sacrifices explanation simplicity. Here, all relevant training examples are used for both prediction and justification, thereby optimizing performance while ensuring fully faithful explanations. However, explaining model predictions via a large number of training examples 
However, explaining model predictions via a large number of training examples can induce information overload, hurting explanation simplicity.


\underline{Case II} achieves high predictive performance and explanation simplicity but sacrifices explanation fidelity. Like Case I, all relevant training examples are used to make the prediction, maximizing performance. However, to simplify the explanation, only a subset of the training examples used to make the prediction is used to explain it. While this simplifies the explanation for the user, it sacrifices explanation fidelity to achieve this. 


Finally, \underline{Case III} sacrifices predictive performance to optimize explanation fidelity and simplicity. In this case, only a subset of relevant training examples is used to make the prediction, reducing performance. However, the same subset used to make the prediction is also used to explain it, yielding a faithful explanation. The virtue of having fewer training examples in the explanation is its simplicity, making it easier to understand.

\section{Wrapper Boxes}
\label{sec:whiteboxes}
\vspace{-0.5em}

Our wrapper box pipeline essentially “wraps” a given neural model with one or more white box classic models fitted on extracted neural representations for inference. Note that the resultant classic models do \textit{not} attempt to approximate the underlying neural model faithfully.
While both classifiers leverage learned linearly separable neural representations, the underlying decision-making process differs. Hence, derived example-based explanations faithfully explain the inference procedure of \textit{the wrapper boxes} (interpretable classic models), \textit{not} the original neural model. 

Post hoc methods evaluate fidelity for the neural model they seek to explain \cite{DeYoung2020-xf, jacovi2020towards} since explanations can diverge from actual model behavior. In contrast, we leverage case-based classifiers where derived example-based explanations by construction must have been consulted during inference. 
Loss in fidelity can only occur intentionally if fewer training examples are shown in the explanation to reduce information overload. 





 
%

\subsection{Learning Feature Representations}


As shown in {\bf Figure~\ref{fig:WrapperBox}}, we start with a fine-tuned neural model that acts as a task-specific encoder to learn high-quality embeddings for the input text. Whereas traditional neural models often fit linear classifiers on learned representations, we extract these representations for use by various classic, white box classifiers. This substitution thus enables prediction supported by faithful, example-based explanations and is agnostic to the neural architecture, training procedure, and data used.

%
%
%

Our only assumption about the neural model is the ability to extract hidden states (or some form of encoded inputs). After training, another pass is made through training data to extract hidden states per token from the penultimate layer. 
%
%
For our sentence-level prediction tasks, we mean pool across tokens to obtain sentence-level representations. Because wrapper boxes rely on feature encodings for prediction, we store them in a format providing fast access: in-memory \href{https://numpy.org/}{\texttt{Numpy}} arrays. 

\subsection{Wrapper Box Models}

We consider three case-based models in which inference is directly linked to training examples. This means that, by design, model predictions can be faithfully and intuitively attributed to specific relevant training examples. 

Building on the conceptual discussion of example-based explanations in Section \ref{sec:explain}, assume the classic model consults $n$ training examples to make a prediction for a given input and that $m <= n$ of these training examples are shown to explain the prediction. When $m<n$ (sacrificing explanation fidelity to boost explanation simplicity), a specific consideration is how each model selects which subset of $m$ examples to show. Intuitively, the $m$ examples should be a representative sample of the complete set of $n$ examples to avoid introducing bias and misleading users \cite{lakkaraju2020fool}. Similarly, when $n$ is reduced (to simplify explanations while preserving $m=n$ explanation fidelity), how to select the smaller subset $n$ of training examples is also model-specific. 

\paragraph{$k$ Nearest Neighbors (\knn)}

\knn predicts the class label for each input according to the dominant class of the $k$ most similar training examples. The nearest neighbors consulted thus constitute faithful, example-based explanations for model predictions. The simplest, unweighted \knn model performs majority voting, whereas weighted \knn weights neighbors by proximity to the input instance. 

\underline{Explanations and Tradeoffs.} \knn uses $n=k$ training examples to make a prediction. 
While we observed relatively small performance differences across the narrow range of $k=n$ values considered above, larger $n$ generally improve predictive performance, while smaller $m$ will simplify explanations. Because \knn inherently orders training examples by proximity to the input, training examples can be easily downsampled, either to make predictions (reduced $n$) or explain them ($m < n$). 

However, when $m<n$ (reducing explanation fidelity to simplify the explanation), 
%
%
%
the majority label of the $m$ nearest neighbors could differ from that of the $n$ nearest neighbors, making the explanation inconsistent with the prediction. In this case, it may be more intuitive to explain the prediction by the $m$ nearest neighbors whose majority label matches that of the $n$ nearest neighbors.

\paragraph{Decision Trees (DTs)}

Decision trees learn a set of rules that act as hyperplanes. Given an input, these rules specify a decision path from the root to a given leaf node. Prediction is based on a majority vote over all training examples assigned to that leaf node. Once constructed, a DT requires the least computation for prediction since decision rules are just simple conditionals. One could even discard all training data after DT construction since only the majority label per leaf node and the final set of rules are needed for inference. However, training data must be kept if we wish to provide example-based explanations \cite{caruana1999case}.

\underline{Explanations and Tradeoffs.} 
Just as \knn labels an input by a majority vote of the $k$ nearest training examples, DT uses a similar vote of the given leaf node's training examples. In both cases, these training instances constitute faithful example-based explanations of the model's prediction.
However, whereas \knn directly selects training examples by similarity to the input, the similarity of leaf node training examples to the input is less direct.   

Because the number of training examples $n$ used to make a prediction (for a given leaf node) may be large, faithfully showing all $m=n$ of the training examples may induce information overload. Just as \knn downsampling would intuitively select the training examples most similar to the input, DT downsampling would also select the most central training examples in the leaf node (to represent the complete leaf set best). When $m < n$ (reducing explanation fidelity to boost simplicity), just as \knn selects the $m$ nearest neighbors whose majority label matches the predicted label, DT selects the $m$ most central training examples whose majority label similarly matches the prediction.

\paragraph{L-Means}

We hypothesize that instances with the same class label may naturally cluster together, assuming a high-quality feature encoding of the domain (such as learned by a fine-tuned DNN). 
%

Inference for \lmeans is the simplest of all wrapper boxes: given an input, we find the closest cluster centroid and assign its label to the input. This reduces the full training set to $L$ representative cluster centroids, which act as rudimentary {\em prototypes} \cite{hase2019interpretable,das2022prototex}. Like DT, inference only requires the majority label of relevant training examples; training data is no longer used once cluster centroids and labels are known. 

\underline{Explanations and Tradeoffs.} As in ProtoTex \cite{das2022prototex}, cluster centroids cannot be directly shown because they are latent. Instead, we must explain model predictions via the training examples that induce each centroid and whose aggregated vote assigns the centroid label. 

Like other models, when the number $n$ of voting training examples is large, showing all $n$ examples can induce information overload. Similar to how DT downsampling selects the most central training examples in the leaf node, $L$-Means downsampling selects the most central training examples in the cluster. When $m < n$ (reducing explanation fidelity to boost simplicity), just as \knn selects the $m$ nearest neighbors whose majority label matches the predicted label, $L$-Means selects the $m$ most central training examples in the cluster whose majority label likewise matches the prediction.

\begin{table*}
\centering
\small
\scalebox{0.99}{
\begin{tabular}{c l c c c c|c c c c|c c c c}
    \toprule
     & & \multicolumn{4}{c|}{\textbf{\bart-large}} &
      \multicolumn{4}{c|}{\textbf{\deberta-large}} & 
      \multicolumn{4}{c}{\textbf{Flan-T5-large}} \\
      & & {Acc.} & {Prec.} & {Rec.} & {F1} & {Acc.} & {Prec.} & {Rec.} & {F1} & {Acc.} & {Prec.} & {Rec.} & {F1} \\
      \cmidrule(r){1-14}
    \multirow{4}{*}{\rotatebox{90}{{ 
        \scalebox{0.8}{%
           \toxigen
        }
    }}} & 
    Original & 80.85 & 67.69 & 70.07 & 68.86 & 82.77 & 70.47 & 73.94 & 72.16 & 81.38 & 72.43 & 61.97 & 66.79 \\
    & KNN & +0.74 & +3.10 & -3.52 & -0.26
        & +0.96 & \textbf{\increase{+5.02}} & \textbf{\decrease{-5.63}} & -0.45
        & 0.00 & -0.71 & +1.41 & +0.50 \\
    & DT  & +0.32 & \textbf{\increase{+5.08}} & \textbf{\decrease{-9.86}} & -2.96
        & -0.22 & \textbf{\increase{+5.62}} & \textbf{\decrease{-11.87}} & -4.07
        & -0.10 & +0.26 & -1.05 & -0.51 \\

    & L-Means & -0.96 & -2.01 & 0.00 & -1.06
    & -0.53 & +0.65 & -4.58 & -1.93
    & -0.21 & -0.04 & -1.06 & -0.64 \\
    \midrule
    \multirow{4}{*}{\rotatebox{90}{{ 
        \scalebox{0.8}{%
           \esnli
        }
    }}} & 
    Original & 90.28 & 90.27 & 90.27 & 90.27 & 91.75 & 91.84 & 91.76 & 91.78 & 90.85 & 90.82 & 90.82 & 90.82 \\
    & KNN & +0.11 & +0.14 & +0.12 & +0.13
        & -0.77 & \textbf{\decrease{-0.84}} & \textbf{\decrease{-0.79}} & \textbf{\decrease{-0.80}} 
        & -0.62 & -0.61 & -0.61 & -0.61 \\ 
    & DT & \textbf{\decrease{-0.92}} & \textbf{\decrease{-0.90}} & \textbf{\decrease{-0.91}} & \textbf{\decrease{-0.91}}
        & +0.18 & +0.11 & +0.17 & +0.16 
        & +0.01 & +0.01 & +0.01 & +0.01 \\
    & L-Means & \textbf{\decrease{-2.82}} & \textbf{\decrease{-1.76}} & \textbf{\decrease{-2.75}} & \textbf{\decrease{-2.61}} 
                 & \textbf{\decrease{-0.84}} & -0.45 & \textbf{\decrease{-0.83}} & -0.77
                 & -0.12 & -0.13 & -0.12 & -0.13 \\
    \bottomrule
  \end{tabular}
}
\vspace{-0.5em}
\caption{\label{tab:Results}
\% change in accuracy (acc.), precision (prec.), recall (rec.), and F1 (macro-averaged) from baseline for wrapper boxes over various transformers, using only representation from the penultimate layer. Statistically significant (see Appendix~\ref{sec:sigtestsProcedure} for procedure) wrapper box results are bolded, with positive results in blue and negative results in red. Table~\ref{tab:BaselineSigTestResults} shows significant differences between the baseline transformers not displayed here.}
  \vspace{-1em}
\end{table*}

\section{Evaluation: Prediction Performance}
\label{sec:exp}

We first compare the predictive performance of wrapper boxes vs.\ underlying neural models. 
Because neural models forecast via linear layers, we expect wrapper boxes to benefit from this learned linear separability and perform comparably. We consider two tasks and datasets:


\vspace{0.5em}
\noindent\textbf{\href{https://huggingface.co/toxigen}{\textsc{\toxigen}}} \citep{hartvigsen2022toxigen} consists of offensive and benign English statements generated by GPT-3 \cite{brown2020language}. We use the 9,900 human-labeled instances, ignoring other instances without gold labels. Each instance is assigned toxicity 
labels on a 5-point scale. We binarize labels by mapping values 1-3 to {\em non-toxic} and 4-5 as {\em toxic} for binary classification.  
%
%
Based on \citeauthor{hartvigsen2022toxigen}'s 90/10 train-test split, we section off a validation set, resulting in a 70/20/10 train-eval-test split. 
The dataset is highly skewed, with a 3:1 ratio of benign vs.\ toxic speech. We employ stratified sampling to maintain this ratio in each split. 

\vspace{0.5em}
\noindent\textbf{\href{https://huggingface.co/datasets/esnli}{\textsc{\esnli}}} \citep{camburu2018snli} adds crowdsourced natural language explanations for the 569,033 English premise-hypothesis pairs originally annotated in SNLI \citep{snli}. We 
follow the predefined training-eval-test splits. 
Each split contains a balanced label distribution. 
Appendix~{\ref{sec:ESNLIQualExs}} compares wrapper box explanations vs.\ those obtained via crowdsourcing. 


\paragraph{Models} We report on three language models: {\bart}-large \citep{lewis2019bart}, {\deberta}-large \citep{he2021debertav3}, and Flan-T5-large \citep{chung2022scaling}, based on checkpoints from \href{https://huggingface.co/}{Huggingface} \citep{wolf2020huggingface}. Representations are extracted from the layer immediately preceding the linear classification head for \bart-large and \deberta-large models. For Flan-T5, representations are extracted from the layer preceding the language generation head. Implementation details for neural and white box models are discussed in Appendix~\ref{sec:AppImpl}.

\begin{table*}[ht]
\centering
\small
\scalebox{0.97}{
\begin{tabular}{llrrrrrr}
\toprule
 &  & \multicolumn{3}{c}{\textsc{Toxigen}} & \multicolumn{3}{c}{\textsc{E-SNLI}} \\
\cmidrule(lr){3-5} \cmidrule(lr){6-8}
 {\bf Classifier} & {\bf Selector} & {$\uparrow$ \bf \!Coverage\%} & {$\uparrow$ \bf \!Correctness\%} & {$\downarrow$ \bf \!Median} & {$\uparrow$ \bf \!Coverage\%} & {$\uparrow$ \bf \!Correctness\%} & {$\downarrow$ \bf \!Median} \\
\midrule
LR & Yang Fast & 27.45 & 27.13 & 51.00 & 89.83 & 0.39 & 76,446.50 \\
LR & Yang Slow & 27.45 & 26.49 & 33.00 & 89.83 & 0.11 & 2.00 \\
\hline
DT & Greedy & 12.02 & 12.02 & 24.00 & 3.23 & 3.23 & 89.00 \\
L-Means & Greedy & 100.00 & 100.00 & 6,377.00 & 100.00 & 100.00 & 140,523.00 \\
KNN & Greedy & 100.00 & 100.00 & 211.00 & 100.00 & 100.00 & 77.50 \\
\bottomrule
\end{tabular}
}
\caption{Benchmarking selectors to derive $S_t$. Coverage is the \% of test inputs for which a $S_t$ was proposed. Correctness is the \% of test inputs for which a $S_t$ was proposed and verified that their removal and retraining led to a prediction flip. Median is the median set cardinality across only the verified subsets that lead to prediction flip.}
\label{tab:SubsetsTable}
\vspace{-1em}
\end{table*}


\subsection{Results}
\label{sec:results}

Results are shown in {\bf Table~\ref{tab:Results}}. Our methodology for significance testing  is described in Appendix \ref{sec:sigtests}. 
%

Wrapper boxes perform largely comparable to baseline transformers for both datasets. For \toxigen, across 48 results per dataset (3 transformers x 4 wrapper boxes x 4 metrics), only 6 of the 48 (12.5\%) differences are statistically significant. For 3 of the 6 cases, the wrapper box performs significantly better than the baseline. For \esnli, while 13 of the 48 ($27$\%) scores show statistically significant differences, whether differences are large enough to be noticeable by users is unclear (Appendix \ref{sec:noticeable}). We observe no significant differences at all with Flan-T5, though note that 
While \deberta is generally the best-performing model. 

Perhaps most remarkable is that the simple \lmeans formulation reduces the entire training set to 2-3 examples that provide the basis for all model predictions, yet still performs competitively.

Appendices~\ref{sec:VisualizingLMeans} and \ref{sec:QualExs} respectively visualize \lmeans clusters and provide  qualitative examples. 

\paragraph{Results for large language models (LLMs)} Appendix~\ref{sec:LLMs} conducts an ablation study that evaluates the effectiveness of wrapper boxes using representations from modern LLMs. Namely, we experiment with Llama 2-7B-Instruct \citep{touvron2023llama}, Llama 3-8B-Instruct \citep{dubey2024llama}, Mistral-7B-Instruct \cite{Jiang2023Mistral7B}, and Gemma-7B-Instruct \cite{Mesnard2024Gemma}. Results show that wrapper boxes using zero-shot LLM representations strongly outperform baseline LLM performance across both tasks. 

\section{Evaluation: Training Data Attribution}
\label{sec:subsets}
\vspace{-0.5em}


The ability to attribute model decisions to specific training data enables decisions to be contested on the basis of the training data responsible. To evaluate how well wrapper boxes support this use case, we adopt \citet{Yang2023-st}'s task formulation of finding a subset of training data $S_t$ that, if removed, would change the model decision for a given input. We use the same two datasets but only with \deberta representations (best performing model). 



\paragraph{Baselines}
\citet{Yang2023-st}'s two algorithms are limited to convex linear classifiers (e.g., logistic regression). We report these as baselines. Appendix~\ref{sec:appendixYangReproduction} details our  reproduction of their reported results, further validating the new results we report with their methods on our own datasets. 

{\bf Yang Fast} (Algorithm 1) uses influence functions to estimate expected change in output probability from removing subset $S_t$. A $S_t$ is only output if the expected change exceeds a threshold $\tau$. 

{\bf Yang Slow} (Algorithm 2) 
starts with all training data and 
seeks to iteratively reduce size of $S_t$ by approximating  expected changes to model parameters $\theta$ upon removal. Like {\em Yang Fast}, $S_t$ is only found if the expected output change exceeds $\tau$.

Of note, \citeauthor{Yang2023-st} report on five binary datasets in their work: three balanced, and two highly skewed 9:1 (``hate'' and ``essays''). While results are strong on the balanced datasets, coverage is low on hate (67\%) and very low on essays (11-12\%). \citeauthor{Yang2023-st}\ remark upon hate's severe label skew, and to address it, select a post hoc $\tau = 0.25$ for this dataset only (using $\tau = 0.5$ for all others). Oddly, they do not note or address the same skew in essays, which may lead the very low coverage reported.





\paragraph{Our Approach} Algorithm~\ref{alg:greedy} defines a greedy approach to derive $S_t$ from wrapper box explanations. For \knn, $C^{\text{tr}}$ includes all neighbors of the input, ranked by proximity. For DT, $C^{\text{tr}}$ comprises all examples in the same leaf, ranked by proximity. For \lmeans, $C^{\text{tr}}$ consists of all points in the same cluster, ranked by proximity to the cluster centroid. Post-filtering, we remove examples in chunks until a prediction flip is observed. $S_t$ is then refined (iteratively or in chunks, depending on $\phi$) until no size reduction is possible. This encourages the derived $S_t$ to be minimal (but still leads to a prediction flip). See Appendices~\ref{sec:Chunking} and ~\ref{sec:KNNAlgo} for further details and an optimized algorithm for \knn (no training).


\begin{algorithm}[t]
\small
\caption{Greedy approach to derive $S_t$ from wrapper box explanations}
\label{alg:greedy}
\hspace*{\algorithmicindent} \textbf{Input:} $f$: Model, $C^{\text{tr}}$: Ranked set of candidate training examples to select from, $x_t$: Test input, $y_t$: Test input label, $B$: Number of bins, $\phi$: Iterative threshold \\
\hspace*{\algorithmicindent} \textbf{Output:} $S_t$, a subset of training points that flips $y_t$ (or $\emptyset$ if unsuccessful) 
\begin{algorithmic}[1]
\Function{FindSubset}{$C^{\text{tr}}, x_t, y_t, B$}
\State $b \gets \lceil \frac{|\mathcal{L}|}{B} \rceil$ \Comment{Bin size}
\State $\mathcal{L} \gets \{(x_i, y_i) \in \mathcal{C^{\text{tr}}} \mid y_i = y_t\}$ \Comment{Filter candidates to match prediction to reduce search complexity}
\For{$i \gets 1$ to $B$}
    \State $C^{\text{tr}}_i \gets C^{\text{tr}} \setminus \{\mathcal{L}[j] \mid j \leq i * b\}$
    \State $\hat{f} \gets \text{train\_model}(C^{\text{tr}}_i)$
    \State $\hat{y}_t \gets \hat{f}(x_t)$
    \If{$\hat{y}_t \neq y_t$}
        \State \Return $\{\mathcal{L}_j \mid j \leq i\}$
    \EndIf
\EndFor
\State \Return $\emptyset$
\EndFunction
\State $S_t \gets$ \Call{FindSubset}{$C^{\text{tr}}, x_t, y_t, B$}
\State $\text{previous\_size} \gets 0$
\While{$|S_t| > 0$ \textbf{and} $|S_t| \neq \text{previous\_size}$}
\State $\text{previous\_size} \gets |S_t|$
\If{$|S_t| < \phi$}
\State $S_t \gets$ \Call{FindSubset}{$S_t, x_t, y_t, |S_t|$}
\Else
\State $S_t \gets$ \Call{FindSubset}{$S_t, x_t, y_t, B$}
\EndIf
\EndWhile
\State \Return $S_t$
\end{algorithmic}
\end{algorithm}

\subsection{Results}


Results in {\bf Table~\ref{tab:SubsetsTable}} report three key metrics: \textit{coverage} (\% of test inputs for which a subset $S_t$ was proposed), \textit{correctness} (\% of test inputs for which removing $S_t$ correctly changed the model decision), and the median size of correct $S_t$ subsets found). 

\textbf{Baselines.} 
\citeauthor{Yang2023-st}'s methods do not perform well. For \toxigen, we suspect the issue is label skew (see discussion above).  
Classifying directly via \deberta vs.\ using logistic regression ($\tau=0.5$) with \deberta representations yielded comparable results (Table~\ref{tab:LogisticRegressionPerformance}), so we use $\tau=0.5$ for \citet{Yang2023-st}'s methods on \toxigen. 




For \esnli, Yang Fast/Slow propose $S_t$ $\sim90$\% of the time, but removing $S_t$ almost never changes model decisions. Because they only consider binary classification tasks, their formulation with $\tau$ likely does not make sense for multi-class tasks like \esnli that typically involve predicting the most probable class through softmax probabilities. 

{\bf Wrapper boxes}. Overall, \knn is the clear winner, with perfect coverage and correctness and far smaller $S_t$ than \lmeans. While both \knn and \lmeans achieve perfect coverage and correctness on both datasets, $S_t$ tends to be quite large for \lmeans since clusters (see Appendix~\ref{sec:VisualizingLMeans}) are mostly homogeneous; many training supporting the model decision must be removed before points with other labels come to the fore to change the decision. 

DT has low coverage because its subset candidate search space is so small, having only leaf examples. This contrasts sharply with \knn (all training examples) and \lmeans (all cluster points). By the same token, when DT does find a $S_t$ subset, it tends to be far smaller than \knn or \lmeans.

\section{Conclusion}

We propose wrapper boxes to provides faithful, example-based explanations for classic case-based model predictions, attributing decisions to specific training data. Our proposed pipeline is quite general and agnostic to the underlying neural architecture, training procedure, and input data. After training a neural model, the learned feature representation is input to white-box case-based reasoning models for prediction. Because case-based models tie inference directly to specific training data, each prediction can be faithfully attributed to the training examples responsible.

Our first evaluation showed that white case-based models could deliver predictive performance largely comparable to baseline transformers, as seen across seven large pre-trained language models, two datasets of varying scale, three classic models, and four metrics.

In addition, we discussed how such attribution enables automated decisions to be contested based on the training data responsible for those decisions. In comparison to prior work \cite{Yang2023-st}, our approach achieves both higher coverage and correctness in identifying which training data to remove to change a model decision. 

Beyond contesting model decisions, other use cases include intuitively explaining decisions to end-users based on past examples or supporting data-centric AI operations for model developers (e.g., training data augmentation and cleaning). 


\section{Limitations}
\label{sec:limitations}


\subsection{Time and Space Requirements}

Wrapper boxes require additional space to store training instances to be presented as example-based explanations. For example, while DT and \lmeans models no longer require training data for inference once trained, they must continue to store training data to provide 
example-based explanations. For DT, 
representative subsets of examples per leaf node may be pre-computed and cached ahead of time for fast explanation retrievals. 
\lmeans is similar: 
since clusters are invariant across all predictions, representative subsets of desired sizes may be pre-computed and cached ahead of time for fast explanation retrievals at inference time. In both cases, storage demands vary depending on the number of desired examples to present for explanations. 

Different wrapper boxes will naturally vary in computation time and space needs, with some models potentially resulting in slower or faster inference than the base neural model. Moreover, we have used relatively simple implementations for each wrapper box. More advanced schemes, e.g., dynamic $k$ for \knn \citep{zhang2017EfficientKNN}, could further increase the computational time or space requirements. Generally, standard computational requirements of classic models are carried forward into our adoption of them as wrapper boxes.

\subsection{Use-Cases for Training Data Attribution} 
\label{sec:no-user-study}

The ability of wrapper boxes to faithfully attribute model decisions to specific training data has a variety of applications. However, our study only evaluates how well wrapper boxes enable model decisions to be contested based on the training data responsible for those decisions. More specifically, we considered the task of 
identifying which training data would need to be removed in order to change a model decision \cite{Yang2023-st} (Section~\ref{sec:subsets}). 

Beyond contesting model decisions, other use-cases include explaining decisions to end-users based on known past examples \citep{Schank2014-bf}. Attribution could also support data-centric AI operations for model developers to help uncover artifacts \citep{lertvittayakumjorn2021explanation}, errors \citep{koh2017understanding},
%
%
and gaps \citep{Khanna2019-ho} in training data, addressed by label cleaning \cite{teso2021interactive,northcutt2021pervasive}
data augmentation \cite{feng2021survey}, and other {\em data-centric} operations \cite{Anik2021-xw}.

Section \ref{sec:RW} noted that prior work has consistently validated the significance, utility, and effectiveness of example-based explanations for users \cite{aamodt1994case, Sormo2005-gx, richter2016case}. 
Benefits include increased model understanding (simulatability), complementary performance, and trust. \cite{Yeh2018-pt, papernot2018deep, Cai2019-kq, hase2020evaluating, han2020explaining, rajagopal2021selfexplain, das2022prototex, suresh2022intuitively, Chen2023-ft}. However, we have yet to actually evaluate any of these benefits in the context of wrapper boxes. 

For both use cases above -- explaining model decisions to end-users or supporting data-centric AI for model developers -- user studies would be valuable to assess the utility of wrapper boxes. 



Of particular interest, Section~\ref{sec:explain} discussed how wrapper boxes permit thoughtful tradeoffs across three key variables of interest: predictive performance, explanation \textit{faithfulness}, and explanation \textit{simplicity}. However, we have yet to investigate these tradeoff space with real users. Future work could conduct user studies to better elucidate how different tradeoffs impact real user experience.

\subsection{\texorpdfstring{$S_t$ Assumptions and Challenges}{S_t Assumptions and Challenges}}

While Section~\ref{sec:subsets} usefully investigates how model decisions can change by counterfactually removing a subset of training data $S_t$, many more counterfactual conditions could be considered that would also alter model decisions, such as the choice of model and training regieme. Neither we nor \citet{Yang2023-st} consider such counterfactual conditions that would be more difficult for end-users to contest. 

Similarly, we and \citeauthor{Yang2023-st} both apply a classification model atop 
fixed feature representations, without considering counterfactual data conditions that would change  feature representations. In \citeauthor{Yang2023-st}'s work, bag-of-words and BERT embeddings are used off-the-shelf and counterfactual training data conditions only impact the learned logistic regression model. In our work, neural representations are fine-tuned on all training data and counterfactual training data conditions only impact wrapper box inference atop neural representations.

As $S_t$ grows, model auditing becomes more difficult, akin to the cognitive overload of showing many examples in a model explanation (Section~\ref{sec:conceptual}). However, prior work \citep{Ilyas2022-hu, Yang2023-st} has shown that large $S_t$ can also indicate predictor robustness, since more training data must be removed to change model decisions. Future work could thus usefully explore tradeoffs of benefits between small vs.\ large $S_t$ subsets.

\section*{Acknowledgments}

This research was supported in part by Cisco and by {\em Good Systems\,}\footnote{\url{http://goodsystems.utexas.edu/}}, a UT Austin Grand Challenge to develop responsible AI technologies. This work was also partially supported by NSF grant IIS-2107524. The statements made herein are solely the opinions of the authors and do not reflect
the views of the sponsoring agencies.

\bibliography{anthology, custom, paperpile}

\clearpage
\appendix


\begin{table}[h]
\centering
\scalebox{0.95}{
\begin{tabular}{c c c c c}
\hline
\textbf{Dataset} & \textbf{Train} & \textbf{Valid} & \textbf{Test} & \textbf{Ratio} \\ \hline
\toxigen & 6980 & 1980 & 940 & 3:1 \\ \hline
\esnli & 549361 & 9842 & 9824 & balanced \\ \hline
\end{tabular}}
\caption{Dataset information. Ratio is the distribution of labels in each split (same for all splits due to stratified sampling).}
\label{tab:dataset_info}
\end{table}

\section{Implementation Details}
\label{sec:AppImpl}

All transformers are fine-tuned using the AdamW optimizer \citep{loshchilov2017decoupled} on \href{https://pytorch.org/docs/stable/generated/torch.nn.CrossEntropyLoss.html}{Cross Entropy loss} over ten epochs, with early stopping \citep{ji2021early} if validation performance degrades for two consecutive evaluations (every 100 steps). 
All layers 
are fine-tuned. We use seed $42$, a learning rate of $1e-5$, and a batch size of $16$ for \bart and \deberta. Since Flan-T5 is larger, we use a learning rate of $1e-4$ and a batch size of $8$. No hyperparameter search is conducted. 
All models are fine-tuned on a single compute node with three NVIDIA A100 GPUs and 256GB of DDR4 RAM within one week of GPU hours. 

Feature encodings are extracted from each trained neural model for use by classic models (i.e., wrapper boxes). We implement Logistic Regression (LR), KNN, and \lmeans using Scikit-Learn {\citep{scikit-learn}}. Early experiments suggest that results were fairly comparable across small values of $k$ (1, 3, 5, 7, and 9) for unweighted and weighted. For this reason, we report unweighted KNN with {{\tt k=5}}. We utilize K-D trees {\citep{BentleyKDTrees}} for efficiently retrieving nearest neighbors. For decision trees, we use \texttt{DecisionTreeClassifier} from Scikit-Learn and set {{\tt max\_depth} = 3} to guard against over-fitting for skewed \toxigen, though this value was not tuned. For \esnli, since the Scikit-Learn implementation does not scale well, we opt for LightGBM {\citep{Ke2017-vl}} with one classifier. For both trees, we set the minimum number of samples in each leaf to be 20. For {\lmeans}, we set {{\tt algorithm='elkan'}} for more efficient computation since our clusters are well-defined. We use $\tau=0.5$ for LR on \toxigen and set \texttt{multi\_class=multinomial} for \esnli. Similarity between data instances in the feature space is measured via Euclidean distance. All results are single-run with random seed 42.

The number of training points and the distribution of labels in each split is shown in Table~\ref{tab:dataset_info}. \toxigen \footnote{\url{https://huggingface.co/toxigen}} is skewed with a 3:1 skew for benign vs.\ toxic examples, respectively. \esnli \footnote{\url{https://huggingface.co/esnli}} is balanced. For \esnli, we excluded 6 training pairs containing blank hypotheses.

\section{Significance Testing}
\label{sec:sigtests}

\subsection{Procedure}
\label{sec:sigtestsProcedure}

We perform statistical testing to A/B test the performance of baseline transformers vs.\ treatment wrapper boxes. We also apply the same procedure to compare baseline transformers to each other. Specifically, correct vs.\ incorrect predictions by each model yield separate binomial distributions. Given relatively large sample sizes, we compute the z-score as shown below \citep{casagrande1978improved}: 

%
\begin{equation}
    z = \frac{\hat{m_1} - \hat{m_2}}{\sqrt{\hat{m}-(1-\hat{m})(\frac{1}{n_1}+\frac{1}{n_2})}}
\end{equation}
where $\hat{m} = \frac{n_1\hat{m_1} + n_2\hat{m_2}}{n_1 + n_2}$. We test the null hypothesis that for some given metric $m$ (e.g. accuracy), there is no significant difference between the two binomial distributions, baseline vs.\ treatment,  
or $m_1 = m_2$ (alternatively $m_1 \neq m_2$). We use $\alpha=0.05$ where results with $p < \alpha$ are significant. We bold and color code significantly different results in Table~\ref{tab:Results}.
Each cell denotes a comparison between a white wrapper box (row) with respect to a baseline transformer (column category) for a particular metric (column type). For transparency, Appendix~\ref{sec:SigTestResults} shows all significance test p-values. \sam{appendix mention here}



\subsection{Ablation Results}
\label{sec:SigTestResults}

{\bf Table~\ref{tab:BaselineSigTestResults}} displays the p-values for all pairwise comparisons between baseline transformers across four metrics (accuracy, precision, recall, F1) on two datasets (\toxigen, \esnli).  {\bf Table~\ref{tab:WrapperBoxSignificanceTests}} shows the p-values for all comparisons between wrapper boxes (rows) and their corresponding baseline transformers (column categories) for the four metrics (columns) on two datasets (row categories).  

\subsection{Are Significant Differences Noticeable?}
\label{sec:noticeable}

In regard to the experimental practice of significance testing, we also wanted to raise a more subtle point here. In system-centered evaluations, we are accustomed to A/B testing of baseline vs.\ treatment conditions and measuring the statistical significance of observed differences. Indeed, we followed this experimental paradigm in this work, showing that wrapper boxes perform largely comparable to that of the underlying neural models.

This experimental paradigm is well-motivated from the standpoint of continual progress, that small but significant differences from individual studies will add up over time to more substantial gains. However, with regard to an individual study, small, statistically significant differences are typically unobservable to users in practice, especially when a slightly less performant system offers some other notable capability, such as providing explanations as well as predictions. 

As a result, an interpretable system that is less performant according to system-based performance metrics may still be experienced as equally performant. \citet{sparck1974automatic} famously remarked that ``statistically significant performance differences may be too small to be of much operational interest'', proposing her classic rule of thumb that only improvements of 5\% or more are \textit{noticeable}, while improvements of 10\% or more are \textit{material}. 

The upshot is that while we are accustomed to placing great weight on minute but statistically significant differences between conditions in system-centered evaluations, from the standpoint of user experience, we should be mindful that such small differences will often be invisible to users, who may well prefer an interpretable system that seems equally performant, even if it actually performs worse by statistical significance testing. 

\begin{table}[h]
\centering
\small
\scalebox{0.95}{
\begin{tabular}{c l c c c c}
    \toprule
    \textbf{Dataset} & \textbf{Classifier} & \textbf{Acc.} & \textbf{Prec.} & \textbf{Rec.} & \textbf{F1} \\
    \midrule
    \multirow{2}{*}{{ 
        \scalebox{1.0}{%
           \toxigen
        }
    }} & Original & 82.77 & 70.47 & 73.94 & 72.16 \\
    & LR & -0.22 & +4.12 & \textbf{-9.86} & -3.23 \\
    \midrule
    \multirow{2}{*}{{ 
        \scalebox{1.0}{%
           \esnli
        }
    }} & Original & 91.75 & 91.84 & 91.76 & 91.78 \\
    & LR & +0.38 & +0.29 & +0.36 & +0.35 \\
    \bottomrule
  \end{tabular}
}
\caption{\% change in accuracy, precision, recall, and F1 (macro-averaged) for logistic regression (LR) using \deberta penultimate representations on \toxigen and \esnli. Like wrapper boxes, logistic regression with \deberta penultimate representations also performs comparably to the original, underlying neural model.}
\label{tab:LogisticRegressionPerformance}
\end{table}

\section{Training Data Attribution Clarifications and Results}
\label{sec:findSubsets}

\subsection{Iterative vs Chunked Removal}
\label{sec:Chunking}

Refitting a new classifier can be expensive, especially for the larger \esnli dataset, when repeated many times. For example, if we were to iteratively remove ranked cluster examples for \lmeans on \esnli, and it takes (empirically) approximately $15$ seconds to retrain and obtain a new test prediction, then finding $S_t$ would take approximately a month on a single node. Hence, in practice, we chunk ranked examples into $B$ consecutive bins such that removal occurs simultaneously for all points in the same bin. Once a $S_t$ is identified this way, we recursively refine iteratively when the subset is less than some iterative threshold  $\phi$, or further split the candidate $S_t$ into smaller $B$ bins in a chunked fashion. Of course, there is likely an efficiency-performance tradeoff here associated with the numbers of bins and $\phi$. As the number of bins increases (smaller chunks), subsets should be minimal but demands more computation. Vice versa, a subset may always be identified (e.g. if the number of bins equals 1, where we are removing the entire candidate set of training points), but it may not be very useful for model auditing. In Section~\ref{sec:subsets}, on both datasets, for DT and \lmeans, we employ $10$ bins (each bin thus consists of 10\% of the training data) and set the iterative threshold to be $\phi = 100$ examples (only do chunk removal when candidate $S_t$ is above this threshold).

To give some qualitative runtimes (on a single node with a 2.1GHz, 48-core Intel Xeon Platinum 8160 "Skylake" CPU) to highlight the infeasibility of iterative removing and retraining for \esnli, Yang fast takes ~3 minutes per example, Yang slow ~15 minutes per example, DT ~5 minutes per example, and \lmeans ~25 minutes per example. \knn is the fastest and finds $S_t$ per example in under 1 second since it requires no retraining (see Algorithm~\ref{alg:knn} detailed in Appendix~\ref{sec:KNNAlgo} below). 

\subsection{Finding Subsets for KNN}
\label{sec:KNNAlgo}

\begin{algorithm}[h]
\caption{Optimized greedy approach to derive $S_t$ for \knn}
\label{alg:knn}
\hspace*{\algorithmicindent} \textbf{Input:} $f$: Model, $C^{\text{tr}}$: Ranked set of candidate training examples to select from, $x_t$: Test input, $y_t$: Test input label  \\
\hspace*{\algorithmicindent} \textbf{Output:} $S_t$, a subset of training points that flips $y_t$ (or $\emptyset$ if unsuccessful) 
\begin{algorithmic}[1]
\State $\mathcal{L} \gets \{(x_i, y_i) \in \mathcal{C^{\text{tr}}} \mid y_i = y_t\}$ \Comment{Filter candidates to match prediction to reduce search complexity}
\For{$i \gets 1$ to $|\mathcal{L}|$}
    \State $C^{\text{tr}}_i \gets C^{\text{tr}} \setminus \{\mathcal{L}[j] \mid j \leq i\}$
    \State $\hat{y}_t \gets \text{majority\_vote} (C^{\text{tr}}_i,k)$ \Comment{Predict using the $k$ nearest neighbors}
    \If{$\hat{y}_t \neq y_t$}
        \State \Return $\{\mathcal{L}_j \mid j \leq i\}$
    \EndIf
\EndFor
\State \Return $\emptyset$
\end{algorithmic}
\end{algorithm}

\knn is a special white box classifier in that there is no "training". The inference module simply remembers the training examples and their labels, while computing nearest neighbors on-demand, given test inputs. When deriving $S_t$, we were thus able to 1) precompute and cache all neighbors and 2) perform iterative "removal" to assess prediction flip without retraining. Specifically, given a test input, we first obtain the ranked list of all neighbors (training points) by proximity. Like algorithm~\ref{alg:greedy}, neighbors are then filtered down to only those with the same label as the prediction as candidates to remove. After filtering, we iteratively remove the nearest neighbor, and then directly examine the next $k$ nearest neighbors to obtain the new prediction. We can do this because our implementation of \knn is unweighted, where it makes predictions by majority vote using the $k$ nearest neighbors. We can thus easily observe if a prediction flip occurred by monitoring if the majority training label in a $k$-sized window has changed without re-training. This makes the greedy approach to identify $S_t$ for \knn the fastest selector amongst all others considered in Section~\ref{sec:subsets}.

\subsection{Extending Our Greedy Selection Algorithm to Influence Functions?}

Algorithm~\ref{alg:greedy} is agnostic to the inference model and the ranking procedure, as long as both are available. From this perspective, one can theoretically leverage influence functions (IF) \citep{koh2017understanding} to obtain training examples ranked by influence to run the procedure. However, because IF assumes linearity and twice-differentiable loss functions, this constrains their application to non-convex white wrapper boxes (e.g., \knn). This restriction is why \citeauthor{Yang2023-st} limit their analysis to logistic regression alone. One could apply IF to the underlying neural module, but it is computationally infeasible to repeatedly retrain the language models analyzed in this study. Our Algorithm~\ref{alg:greedy} is feasible in our experiments because our wrapper boxes are lightweight,  e.g., \knn does not require retraining!  

\subsection{Comparisons to Prior Work}

Algorithm~\ref{alg:greedy} is similar to {\em data models} \citep{Ilyas2022-hu} in that some model retraining is required. However, we note several distinctions here. First, data models are {\em surrogate models} trained to approximate the output of a black-box neural model. Thus, subsets identified through data models are not guaranteed to lead to a prediction flip, whereas we are guaranteed that resultant subsets are correct. Second, learning data models requires collecting labels (probability outputs) from the neural module retrained with different subsets of the training set. This is considerably more expensive since we only retrain lightweight white box models. Third, data models are affected by the stochastic nature of the neural model and its supervised learning framework, so its outputs are nondeterministic and only approximate the neural. On the other hand, our approach yields deterministic subsets, since the re-trained wrapper model must have the same hyperparameters as the original. 

Overall, this paper and prior work \citep{Ilyas2022-hu, Yang2023-st} have shown that finding $S_t$ is computationally challenging. Our greedy approach (besides \knn) requires retraining a new white box classifier for each removal, \citet{Yang2023-st} necessitates inverse hessian approximations, and \citet{Ilyas2022-hu} requires numerous retraining of models to obtain labels for data models. Despite these computational demands, none of these approaches guarantee that identified subsets are minimal (smallest possible) or unique (for each test input).  An alternative direction, as investigated in \citep{Yang2024-tb}, is to flip training labels instead of removing whole examples. 
While this method is sample-efficient for binary classification tasks, its efficacy in multiclass tasks remains uncertain. 

\subsection{Reproducing Yang et al. (2023)}
\label{sec:appendixYangReproduction}

\begin{table*}[h]
\centering
\scalebox{1.0}{
\begin{tabular}{c c c c c}
    \toprule
    \textbf{Dataset} & \textbf{Selector} & \textbf{Coverage} & \textbf{Correctness} & \textbf{Median} \\
    \midrule
    \multirow{2}{*}{{ 
        \scalebox{1.0}{%
           Movie reviews
        }
    }} & Yang Fast & 64.22 & 64.11 & 94 \\
    & Yang Slow & 64.22 & 63.19 & 76 \\
    \midrule
    \multirow{2}{*}{{ 
        \scalebox{1.0}{%
           Essays
        }
    }} & Yang Fast & 7.47 & 7.47 & 24 \\
    & Yang Slow & 7.47 & 7.16 & 12 \\
    \midrule
    \multirow{2}{*}{{ 
        \scalebox{1.0}{%
           Emotion
        }
    }} & Yang Fast & 71.78 & 71.78 & 64 \\
    & Yang Slow & 71.78 & 70.79 & 51 \\
    \midrule
    \multirow{2}{*}{{ 
        \scalebox{1.0}{%
           Hate speech
        }
    }} & Yang Fast & 52.94 & 52.66 & 135 \\
    & Yang Slow & 46.41 & 44.16 & 103 \\
    \midrule
    \multirow{2}{*}{{ 
        \scalebox{1.0}{%
           Tweet Sentiment
        }
    }} & Yang Fast & 89.80 & 89.50 & 110 \\
    & Yang Slow & 75.30 & 60.30 & 213 \\
    \bottomrule
  \end{tabular}
}
\caption{Coverage, Correctness, and Median for logistic regression using BERT \texttt{[cls]} representations for the two $S_t$ selector algorithms proposed in \citet{Yang2023-st}. Note that results differ slightly from their Table 2 due to stochastic differences in generating dataset splits and potential differences in L2 penalty term.}
\label{tab:YangReproductionSubsets}
\end{table*}

Our results in Section~\ref{sec:subsets} show that both selectors from \citet{Yang2023-st} perform poorly on our two selected datasets. To demonstrate that the baseline was implemented correctly (and thereby validate our baseline results with \citeauthor{Yang2023-st}'s methods in our own experiments), we reproduce reported results on the datasets evaluated in \citet{Yang2023-st}, including  Movie sentiment\footnote{\scriptsize \url{https://github.com/successar/instance_attributions_NLP/tree/master/Datasets/SST}} \citep{Socher2013-lx}; Twitter sentiment classification\footnote{\scriptsize \url{https://www.kaggle.com/datasets/kazanova/sentiment140}} \citep{Go2009-nv}; Essay grading\footnote{\scriptsize\url{https://www.kaggle.com/competitions/asap-aes/data}} \citep{Essay2012Hamner}; Emotion classification\footnote{\scriptsize\url{https://huggingface.co/datasets/dair-ai/emotion}} \citep{Saravia2018-qi}, and; Hate speech detection\footnote{\scriptsize\url{https://huggingface.co/datasets/odegiber/hate_speech18}} \citep{De_Gibert2018-tf}. We adopt their source code\footnote{\scriptsize\url{https://github.com/ecielyang/Smallest_set}} to reproduce their results.

As part of reproduction work, we share a few data-cleaning details not reported in \citet{Yang2023-st}. Although their training and testing split sizes are provided in their Table A1, the random seed used to generate those splits was unavailable (besides movie reviews, which appear to use the provided train split and the validation split for testing). Consequently, we observe slightly different subset results than those reported in their Table~2. 

To preserve the split sizes reported in their Table A1, we use the the movie review dataset training split as-is while using the provided validation set as-is for testing. We only use the provided training set for all other datasets and break it down to a 90/10 train-test split. For essays, we first binarize the training split dataset by converting the top 10\% of essay scores to 1 and the rest to 0. Only examples labeled with "sadness" (0) and "joy" (1) were kept for emotions. For hate speech, we similarly only kept training examples labeled with "nohate" (0) and "hate" (1). 19,000 random training points were sampled from the tweet sentiment training split. 

Representations using the \texttt{[cls]} token of \texttt{bert\_base\_uncased}\footnote{\scriptsize\url{https://huggingface.co/google-bert/bert-base-uncased}} \citep{devlin2018bert} were extracted for each dataset, following \citep{Yang2023-st}. For recency, we did not reproduce results with bag-of-words embeddings. We then fitted a logistic regression for each dataset using the extracted BERT representations, using $\tau = 0.25$ for hate speech and $\tau = 0.5$ for all other datasets as specified in \citep{Yang2023-st}. 


Table~\ref{tab:YangReproductionSubsets} shows the subset results for various classifiers and selector methods on the five datasets. We apply the same metrics as described in Table~\ref{tab:SubsetsTable}, noting that Coverage and Correctness are equivalent to the columns "Found $S_t$" and "Flip Successful" in \citet{Yang2023-st}'s Table 2. 

Our reproduced results are comparable to their reported results, barring stochastic differences due to different train/test splits and potentially different L2 penalty terms for each fitted regressor. Furthermore, as remarked in their limitations, "assuming a stochastic parameter estimation method (e.g., SGD) the composition of $S_t$ may depend on the arbitrary random seed, similarly complicating the interpretation of such sets," so it is not surprising that we observe somewhat different outcomes. 


\section{AI vs.\ Human Perceptions of Similarity}
\label{sec:perception}

In explaining model decisions to end-users by attributing decisions to specific training data, we assume that users will understand why the given training examples shown are relevant to a given input.
Otherwise, the training examples shown could appear spurious, and users might not understand why the model deemed these training examples relevant to the input at hand. 
This raises two key questions: 1) how do wrapper boxes measure instance similarity, and 2) how closely does this measurement align with human perceptions of instance similarity?

Given the neural model's extracted embeddings, wrapper boxes compute similarity between instances via Euclidean distance. Instance similarity is thus assessed as a combination of the embedding space and the distance function. 

Human perceptual judgments may naturally diverge from large pre-trained language models' learned latent representation space. For example, \citet{liu2023learning} show that nearest neighbor images using ResNet \citep{he2016deep} representations may not align with human similarity judgments. However, they also show that more human-aligned representations can be learned to improve human decision-making. Similarly, \citet{toneva2019interpreting} show that modifying BERT \citep{devlin2018bert} to match human brain recordings better enhances model performance. 

We hypothesize that continuing progress in developing increasingly powerful pre-trained large language models will naturally trend toward producing representations having greater alignment with human perceptions \citep{muttenthaler2023human}. Of course, it is also possible for more performant embeddings to diverge from human perceptions of similarity. As discussed, alignment between model vs.\ human embeddings can also be directly optimized \cite{liu2023learning,toneva2019interpreting}. 

Of course, this, too, has a risk: tuning embeddings for perceptual judgments could improve explanation quality for users but reduce performance. Thus, the choice of embeddings may embody a tradeoff between performance and explanation quality, though current evidence suggests otherwise \citep{liu2023learning, muttenthaler2023human}.

\section{When are Example-based Explanations Most Appropriate to Use?}
\label{sec:DataTypes}

Different input formats (e.g., text versus image) or categories (e.g., tweets vs. passages) impose varying amounts of cognitive load required to process and reason about analogical justifications. For example, the amount of critical thinking necessary to comprehend social media posts compared to scientific papers is drastically disparate, and that difference may even vary amongst users. 

Consequently, user cognitive load in understanding example-based explanations is likely positively correlated with the amount of information inherently embedded in the inputs themselves. This directly impacts our analysis of explanation simplicity. For tweets, perhaps three or five short posts are still manageable. For scientific passages, maybe even one manuscript is overwhelming. 

If model explanations are intended to support people, quantifying the degree to which explanations actually improve human performance in practice will ultimately require user studies. 

\begin{table*}
\centering
\scalebox{1.0}{
\begin{tabular}{c c c c c}
    \toprule
    & Metric & \bart vs. \deberta & \bart vs. Flan-T5 & \deberta vs. Flan-T5 \\
    \midrule
    \multirow{4}{*}{\rotatebox{90}{{ 
        \scalebox{1.0}{%
           \toxigen
        }
    }}} 
    & Accuracy & 0.282 & 0.768 & 0.434 \\
    & Precision & 0.192 & \textbf{0.0248} & 0.347 \\
    & Recall & 0.0614 & \textbf{<1e-3} & \textbf{<1e-3} \\
    & F1 & 0.116 & 0.338 & \textbf{0.0114} \\
    \midrule
    \multirow{4}{*}{\rotatebox{90}{{
        \scalebox{1.0}{%
           \esnli
        }
    }}} 
    & Accuracy & \textbf{<1e-3} & 0.171 & \textbf{0.024} \\
    & Precision & \textbf{<1e-3} & 0.188 & \textbf{0.0112} \\
    & Recall & \textbf{<1e-3} & 0.180 & \textbf{0.0203} \\
    & F1 & \textbf{<1e-3} & 0.187 & \textbf{0.0168} \\
    \bottomrule
\end{tabular}}
\caption{\label{tab:BaselineSigTestResults} \toxigen and \esnli test set p-values for comparisons between baseline transformers in Table~\ref{tab:Results}}
\end{table*}


\begin{table*}[p]
\centering
\scalebox{0.8}{
\begin{tabular}{c l c c c c c c c c c c c c c}
    \toprule
    & & \multicolumn{4}{c}{\textbf{\bart-large}} &
      \multicolumn{4}{c}{\textbf{\deberta-large}} & 
      \multicolumn{4}{c}{\textbf{Flan-T5-large}} \\
    & & {Acc.} & {Prec.} & {Rec.} & {F1} & {Acc.} & {Prec.} & {Rec.} & {F1} & {Acc.} & {Prec.} & {Rec.} & {F1} \\
    \cmidrule(r){1-14}
    \multirow{3}{*}{\rotatebox{90}{{ 
        \scalebox{0.75}{%
           \toxigen
        }
    }}}
    & KNN & 0.679 & 0.145 & 0.101 & 0.905 & 0.578 
        & \textbf{0.014} & \textbf{0.007} & 0.830 & 1.000 & 0.730 & 0.528 & 0.819 \\
    & DT & 0.860 & \textbf{0.016} & \textbf{<1e-3} & 0.171 
        &  0.903 & \textbf{0.006} & \textbf{<1e-3} & 0.054 & 0.953 & 0.899 & 0.638 & 0.815 \\
    & L-Means & 0.601 & 0.355 & 1.000 & 0.623 & 0.762 
        & 0.757 & 0.028 & 0.355 & 0.906 & 0.983 & 0.638 & 0.770 \\
    \midrule
    \multirow{3}{*}{\rotatebox{90}{{ 
        \scalebox{0.75}{%
           \esnli
        }
    }}}
    & KNN & 0.791 & 0.740 & 0.777 & 0.762 & 0.054 
        & \textbf{0.036} & \textbf{0.049} & \textbf{0.047} & 0.137 & 0.142 & 0.144 & 0.145 \\
    & DT & \textbf{<1e-3} & \textbf{<1e-3} & \textbf{<1e-3} 
        &  \textbf{<1e-3} & 0.796 & 0.762 & 0.758 & 0.774 & 0.843 & 0.831 & 0.841 & 0.832 \\
    & L-Means & \textbf{<1e-3} & \textbf{<1e-3} 
        & \textbf{<1e-3} & \textbf{<1e-3} & \textbf{0.035} & 0.259 & \textbf{0.039} & 0.053 & 0.767 & 0.755 & 0.764 & 0.756 \\
    \bottomrule
  \end{tabular}
}
\caption{\label{tab:WrapperBoxSignificanceTests}\toxigen and \esnli test set p-values for Table~\ref{tab:Results}.}
\end{table*}



\begin{figure*}
    \centering
    \includegraphics[width=16cm]{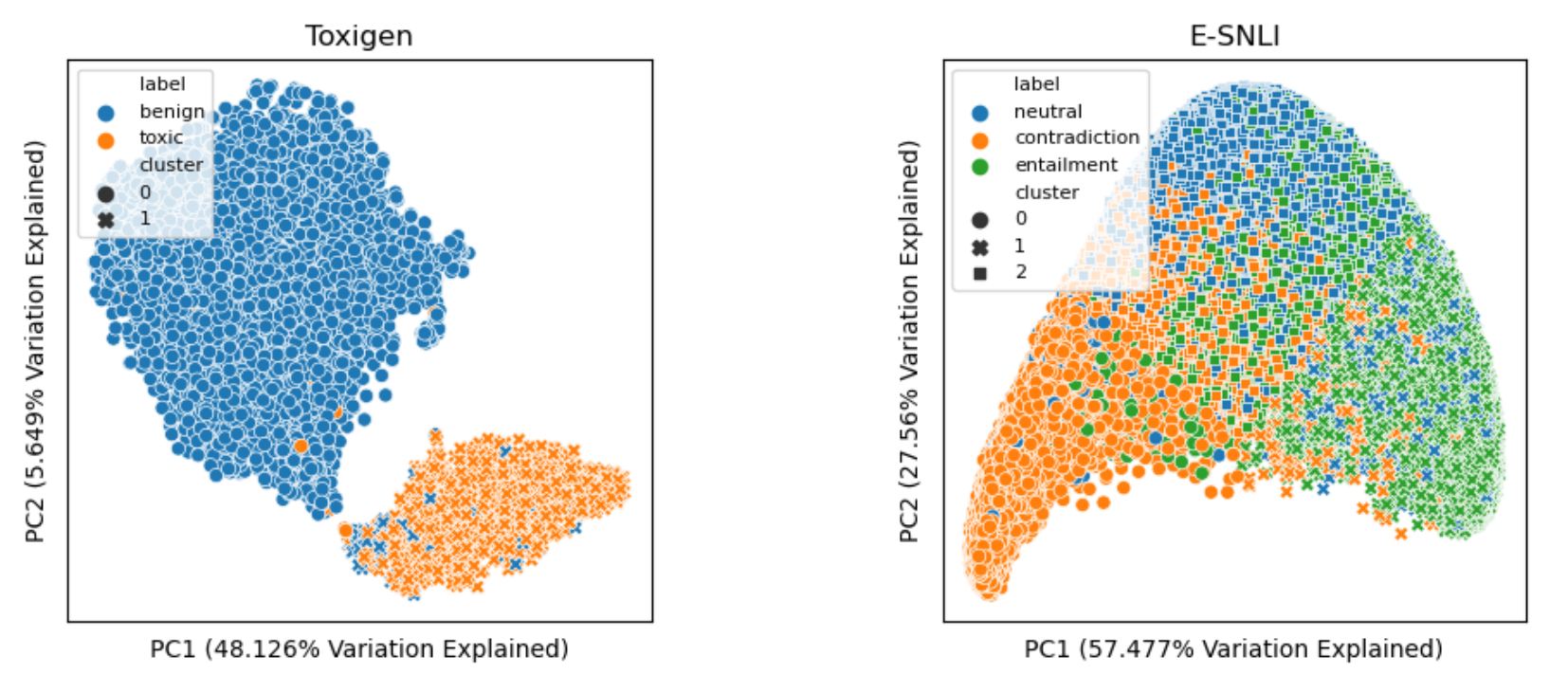}
    \caption{Visualization of resultant clusters from L-Means after PCA with two components.}
    \label{fig:LMeans}
\end{figure*}

\section{Visualizing L-Means}
\label{sec:VisualizingLMeans}

{\bf Figure~\ref{fig:LMeans}} visualizes $L$-Means (described in Section~\ref{sec:whiteboxes}) clusters for the training split of \toxigen and \esnli after using Principal Components Analysis \citep{mackiewicz1993principal} to reduce dimensions of extracted representations. Given limited space, only representations from \deberta-large, the top-performing model, are used. 

\subsection{\toxigen}

For \toxigen, we observe an almost clean clustering of examples. Specifically, cluster 1 (crosses) consists mostly of toxic examples (1671 orange), with only 129 benign (blue) instances. Similarly, cluster 0 (circles) consists mostly of benign (5154 blue) examples, with only 26 toxic (orange) examples. As expected, the two clusters mainly separate along the first principal component, which  accounts for almost half of the variation in \deberta encodings for \toxigen. 

\subsection{\esnli}

We observe that the resultant clusters for \esnli are not as clean as those for \toxigen, hence resulting in noisier predictions that lead to slightly worse performance as shown in Table~\ref{tab:Results}. This may be attributed to the fact that both principal components (PCs) constitute a significant amount of variation for \esnli, whereas PC1 for \toxigen is the sole dominant axis. Specifically, cluster 0 (circles) consists of 1168 entailment, 3996 neural, and 164217 contradiction examples. Cluster 1 contains 163618 entailment, 14229 neural, and 1437 contradiction examples. Cluster 2 comprises of 8628 entailment, 164537 neural, and 17531 contradiction examples.



\begin{table*}[tbp]
\centering
\scalebox{0.8}{
    \begin{tabular}{p{0.4\linewidth}|c c p{0.5\linewidth}}
        \toprule
        Input & Classifier & Prediction &  Explanation \\
        \midrule
        people who are mentally disabled should be treated with the same respect and dignity that we would give to any other human being [Benign]
         & \knn & Benign & people who are mentally disabled should be treated with the same respect and dignity as anyone else [Benign] \\
        & DT & Benign & Same as KNN \\
        & \lmeans & Benign & Same as KNN \\
        \midrule

        $<$GROUP$>$ are very hostile people [Harmful]
        & \knn & Harmful & $<$GROUP$>$ people arent normal [Harmful] \\\\
        & DT & Harmful & Same as KNN \\
        & \lmeans & Harmful & Same as KNN \\
        \midrule

        the government should not be in the business of providing services for those who cannot help themselves [Harmful]
        & \knn & Benign & society should not be obligated to provide for the needs of those who are not able to provide for themselves [Harmful] \\\\
        & DT & Harmful & it is a waste of money to provide services for those who are not productive members of the community [Harmful] \\\\
        & \lmeans & Harmful & Same as KNN \\
        \bottomrule
      \end{tabular}
    }
  \caption{\label{tab:ToxigenExamples}Qualitative examples for \toxigen validation set. Often, \knn, DT, and \lmeans identify the same closest training point as explanations. However, sometimes (row 3) wrapper boxes can lead to diverse explanations, and the final prediction may not agree with the label of the closest training point, since these explanations are parsimonious but not faithful. Nevertheless, we believe that these explanations are still useful, serving as intuitions for correct predictions and clarifications for failure cases.}
\end{table*}


\begin{table*}[tbp]
\centering
\scalebox{0.8}{
    \begin{tabular}{p{0.4\linewidth}|c c p{0.5\linewidth}}
        \toprule
        Input & Classifier & Prediction & Explanation \\
        \midrule

        \multirow{3}{=}{\premise Two women are embracing while holding to go packages. \newline\newline
        \hypothesis Two woman are holding packages.
        [Entailment] \newline\newline
        \human Saying the two women are holding packages is a way to paraphrase that the packages they are holding are to go packages.}
        & \knn & Entailment & 
        \premise Two boys show off their stained,  blue tongues. \newline
        \hypothesis boys are showing their tongues. \newline [Entailment]
        \\\\
        & DT & Entailment & Same as KNN \\\\
        & \lmeans & Entailment & 
        \premise This young child is having fun on their first downhill sled ride. \newline
        \hypothesis A child on a sled. [Entailment]  \\
        \midrule

        \multirow{3}{=}{\premise A shirtless man is singing into a microphone while a woman next to him plays an accordion. \newline\newline
        \hypothesis He is playing a saxophone.
        [Contradiction] \newline\newline
        \human A person cannot be singing and playing a saxophone simultaneously.}
        & \knn & Contradiction & 
        \premise A woman is sitting on a steps outdoors playing an accordion. \newline
        \hypothesis Someone is playing a piano. [Contradiction]
        \\\\
        & DT & Contradiction & Same as KNN \\\\
        & \lmeans & Contradiction & 
        \premise Africans gather water at an outdoor tap. \newline
        \hypothesis Africans are gathering rice for a meal.  [Contradiction]  \\
        \midrule

        \multirow{3}{=}{\premise A woman in a gray shirt working on papers at her desk. \newline\newline
        \hypothesis Young lady busy with her work in office.
        [Neutral] \newline\newline
        \human All women are not young. Although she is working on papers at her desk, it does not mean that she is busy or that she's in an office.}
        & \knn & Neutral & 
        \premise Man raising young boy into the clear blue sky. \newline
        \hypothesis Father holds his son in the air. [Entailment]
        \\\\
        & DT & Entailment & \premise Two soccer players race each other during a match while the crowd excitedly cheers on. \newline
        \hypothesis Two men compete to see who is faster during soccer. [Entailment] \\\\
        & \lmeans & Neutral & 
        \premise A model poses for a photo shoot inside a luxurious setting. \newline
        \hypothesis a woman poses. [Neutral] \\
        \bottomrule
      \end{tabular}
    }
  \caption{\label{tab:ESNLIExamples}Qualitative examples for \esnli validation set. Here, most of the time, \knn and DT identify the same closest training point as explanations, and \lmeans pinpoints a different but related instance (rows 1-2). We postulate that this differs from \toxigen because \lmeans results in less clean clusters.
  Again, sometimes (row 3) wrapper boxes can still lead to diverse explanations, and the final prediction may not agree with the label of the closest training point. Nevertheless, we believe that these explanations are still useful as rationales behind annotated human explanations often also apply to the selected training examples.}
\end{table*}


\section{Qualitative Examples}
\label{sec:QualExs}

We closely examine some qualitative example-based explanations for \deberta-large on the validation splits of \toxigen and \esnli. For the sake of space, we present the single closest (as measured by Euclidean distance over the latent representation space) neighbor/support vector/leaf node instance/cluster point as example-based explanations for various wrapper boxes. Instances are presented as is without any modifications, except that target groups and countries of offensive examples are demarcated in angle brackets. Raw examples may contain punctual, spelling, or grammatical errors. Labels for each training instance are in normal brackets.

Only presenting the closest training point does not constitute a faithful explanation of the wrapper boxes' reasoning process. Here, we again rely on the assumption that relevant examples as judged by the wrapper boxes will also be judged as related by users. Following this logic, we think the closest training point as deemed by the model is likely a good analogy to the input example from the users' perspective as well. However, the closest training point may not be representative of the overall distribution of the training instances applied for inference (e.g. row 3 in Table~\ref{tab:ToxigenExamples}). Even if it may be relevant to the input, the closest example could be an outlier or an atypical example that does not accurately represent the majority of examples employed for reasoning. If users believe that explanations are faithful when they are not, this misinterpretation may also trick users into trusting faulty models \citep{lakkaraju2020fool}.

\subsection{\toxigen}

{\bf Table~\ref{tab:ToxigenExamples}} showcases qualitative examples for \toxigen. Although not faithful, we observe that these explanations are relevant to the input text and are often identical across wrapper boxes. Specifically, \knn, DT, and \lmeans usually pinpoint the same training instances as explanations. 
Rows 1-2 illustrate this phenomenon, where all example-based explanations address the same topic as the inputs. 


Since all wrapper boxes leverage more than just the closest training example in inference (Section~\ref{sec:whiteboxes}), these explanations are simple but are not faithful (Case II in Table~\ref{tab:Strategies}). This can lead to scenarios (row 3) where the final prediction disagrees with the explanation label. Furthermore, there's no guarantee that \knn, DT, and \lmeans always pinpoint the same explanations, and indeed they can be different since the exact mechanism by which similar examples are identified for each approach varies. Either way, we theorize that these explanations are useful to cultivate intuitions for correct predictions and clarifications for failure cases. 

\subsection{\esnli}
\label{sec:ESNLIQualExs}

{\bf Table~\ref{tab:ESNLIExamples}} presents qualitative examples for \esnli. Each sample constitutes a premise-hypothesis pair, alongside a randomly selected (from three) human-annotated explanation. Although our qualitative example-based explanations ( Table~\ref{tab:ToxigenExamples} and Table~\ref{tab:ESNLIExamples}) are simple and intuitive, other NLP tasks may differ, such as topic modeling or passage retrieval. 

Interestingly, whereas for \toxigen \knn, DT, and \lmeans often pinpoint the same training instances as explanations, for \esnli instead \knn and DT follow this trend (rows 2-3). We postulate that this occurs because \lmeans results in less clean clusters (noisier neighbors)


Nevertheless, we observe that provided example-based explanations often require the same reasoning skills as the input, consistent with examples from \toxigen. Again, sometimes (row 3) wrapper boxes can still lead to different explanations, and the final prediction may not agree with the label of the closest training point. 

Unfaithful explanations can still be useful. For \esnli, we observe that rationales behind annotated human explanations often apply to the presented example-based explanations. For example, the human justifies the pair as entailment for the first input example (row 1) since the hypothesis paraphrases the premise. Likewise, our example-based explanation displays a hypothesis that paraphrases the premise.

\section{Wrapper Box Results for LLMs}
\label{sec:LLMs}

\subsection{Evaluation Setup}
We additionally experiment with several more modern large language models (LLMs) to test the generalizability of wrapper boxes. Namely, we experiment with Llama 2-7B Instruct \citep{touvron2023llama}, Llama 3-8B-Instruct \citep{dubey2024llama}, Mistral-7B Instruct \cite{Jiang2023Mistral7B}, and Gemma-7B Instruct \cite{Mesnard2024Gemma}. We used model checkpoints publicly hosted on Hugging Face. Table~\ref{tab:LLM_prompts} shows the prompts used for all LLMs.

Representations are extracted from the penultimate layer (directly preceding the language modeling heads) and mean-pooled across tokens to obtain a sentence-level embedding for white classic models. Sentence representations are then further processed by fitting a logistic regression, and we then take the logit output of the regression model to be the final input features for wrapper boxes. We empirically observe that this logistic transformation is necessary to achieve comparable performance and that fitting wrapper boxes directly on mean-pooled embeddings can lead to severe performance degradation, particularly for \lmeans. 

We use the same datasets and metrics introduced in Section~\ref{sec:exp}. Due to time and computation constraints, we only report zero-shot results and only report metrics on a 10,000 random stratified sample for \esnli. Applying state-of-the-art in-context-learning (ICL) strategies or fine-tuning may improve baseline neural performance but would also result in different representations as input to wrapper boxes. The efficacy of wrapper boxes in these scenarios would thus need to be empirically benchmarked, although we anticipate that there should be no drastic performance degradation. 

\subsection{\toxigen LLM Results}

Table~\ref{tab:LLM_results_toxigen_with_logistic_transformation} shows predictive performance results for \toxigen. LLM zero-shot results favor recall over precision, likely due to the imbalanced distribution of labels in \toxigen, with Llama 3 8B being the best model. No wrapper box uniformly outperforms others, although the precision and recall scores are more balanced. Results show that wrapper boxes using zero-shot LLM representations strongly outperform baseline LLM performance across both tasks. 


\begin{table*}[ht]
\centering
\begin{tabular}{p{3cm}ccccc}
\hline
Model & Classifier & Accuracy (\%) & F1 Score (\%) & Precision (\%) & Recall (\%) \\
\hline
\multirow{5}{*}{LLama2 7B} 
& Zero-shot & 61.17 & 55.76 & 42.51 & \textbf{80.99} \\
& KNN & 82.02 & 68.05 & \textbf{73.47} & 63.38 \\
& DT & 80.85 & 69.90 & 66.56 & 73.59 \\
& L-Means & \textbf{82.23} & \textbf{71.06} & 69.97 & 72.18 \\
\hline
\multirow{5}{*}{LLama3 8B} 
 & Zero-shot & 71.70 & \textbf{65.81} & 51.82 & \textbf{90.14} \\
 & KNN & \textbf{77.23} & 59.47 & \textbf{64.34} & 55.28 \\
 & DT & 76.70 & 61.10 & 61.65 & 60.56 \\
 & L-Means & 76.60 & 61.27 & 61.27 & 61.27 \\
\hline
\multirow{5}{*}{Mistral 7B} 
 & Zero-shot & 68.62 & \textbf{64.50} & 48.99 & \textbf{94.37} \\
 & KNN & 78.72 & 62.69 & 66.67 & 59.15 \\
 & DT & \textbf{79.04} & 63.59 & \textbf{66.93} & 60.56 \\
 & L-Means & 78.72 & 63.90 & 65.56 & 62.32 \\
\hline
\multirow{5}{*}{Gemma 7B} 
 & Zero-shot & 55.64 & 56.79 & 40.23 & \textbf{96.48} \\
 & KNN & 79.36 & 63.12 & \textbf{68.60} & 58.45 \\
 & DT & 78.94 & \textbf{68.37} & 62.57 & 75.35 \\
 & L-Means & \textbf{79.47} & 66.55 & 65.53 & 67.61 \\
\hline
\end{tabular}
\caption{\toxigen test set performance metrics (percentage) of wrapper boxes for various instruction-tuned, open-source large language models. Embeddings are mean-pooled overall all sequence tokens from the penultimate layer (directly preceding the generative language modeling head). Then, representations are further transformed into logits using weights learned from logistic regressor before input as features to other wrapper boxes (KNN, DT, L-Means).}
\label{tab:LLM_results_toxigen_with_logistic_transformation}
\end{table*}

\subsection{\esnli LLM Results}

Table~\ref{tab:LLM_results_esnli_post_logistic_transform} shows predictive performance results for \esnli. Precision and recall scores are balanced for both LLMs and wrapper boxes here since \esnli has a balanced distribution of labels. Interestingly, we observe that decision tree consistently outperforms other wrapper boxes, although the differences (e.g., compared to KNN) may not be significant. Nevertheless, we consistently observe that wrapper boxes can strongly outperform baseline LLMs using zero-shot LLM representations. 

\begin{table*}[ht]
\centering
\begin{tabular}{p{3cm}ccccc}
\hline
Model & Classifier & Accuracy (\%) & F1 Score (\%) & Precision (\%) & Recall (\%) \\
\hline
\multirow{5}{*}{LLama2 7B} 
 & Zero-shot & 47.96 & 42.28 & 50.52 & 48.51 \\
 & KNN & 70.80 & 70.76 & 70.92 & 70.72 \\
 & DT & \textbf{73.03} & \textbf{72.91} & \textbf{72.91} & \textbf{72.96} \\
 & L-Means & 65.95 & 66.42 & 69.37 & 65.87 \\
\hline
\multirow{5}{*}{LLama3 8B}
 & Zero-shot & 61.93 & 62.71 & 68.48 & 61.89 \\
 & KNN & 70.75 & 70.72 & 70.86 & 70.67 \\
 & DT & \textbf{73.35} & \textbf{73.31} & \textbf{73.34} & \textbf{73.30} \\
 & L-Means & 66.43 & 66.83 & 69.53 & 66.33 \\
\hline
\multirow{5}{*}{Mistral 7B} 
 & Zero-shot & 70.99 & 71.47 & 72.81 & 70.99 \\
 & KNN & 80.23 & 80.15 & 80.13 & 80.18 \\
 & DT & \textbf{81.94} & \textbf{81.90} & \textbf{81.91} & \textbf{81.91} \\
 & L-Means & 77.20 & 76.27 & 76.59 & 77.00 \\
\hline
\multirow{5}{*}{Gemma 7B} 
 & Zero-shot & 52.56 & 48.12 & 70.30 & 53.20 \\
 & KNN & 66.07 & 65.98 & 66.15 & 65.97 \\
 & DT & \textbf{69.68} & \textbf{69.68} & \textbf{69.80} & \textbf{69.62} \\
 & L-Means & 62.61 & 63.02 & 66.41 & 62.50 \\
 \hline
\end{tabular}
\caption{\esnli test set performance metrics (percentage) of wrapper boxes for various instruction-tuned, open-source large language models. Embeddings are mean-pooled overall all sequence tokens from the penultimate layer (directly preceding the generative language modeling head). Then, representations are further transformed into logits using weights learned from logistic regressor before input as features to other wrapper boxes (e.g., KNN, DT, L-Means). The training set is a 10,000 random stratified sample of the original ~500k set.}
\label{tab:LLM_results_esnli_post_logistic_transform}
\end{table*}

\begin{table}[h!]
    \centering
    \begin{tabular}{ll}
        \toprule
        \textbf{Dataset} & \textbf{Prompt} \\
        \midrule
        \textbf{\toxigen} & \parbox{0.8\textwidth}{You are a content moderator tasked with toxicity detection. After reading the provided text, categorize it as either "0" for benign or "1" for toxic. Only respond with "0" or "1". Do not include any additional text or explanations.} \\
        \midrule
        \textbf{\esnli} & \parbox{0.8\textwidth}{You are an informative AI assistant tasked with natural language inference. Given the following premise and hypothesis, classify their relationship as "0" for entailment, "1" for neutral, or "2" for contradiction. \\ 
        \textbf{Entailment (0)}: The hypothesis logically follows from the premise. If the premise is true, the hypothesis must also be true. \\
        \textbf{Neutral (1)}: The hypothesis is neither definitively true nor false based on the premise. The premise provides some information, but it is insufficient to confirm or deny the hypothesis. \\
        \textbf{Contradiction (2)}: The hypothesis directly conflicts with the premise. If the premise is true, the hypothesis must be false. \\
        Only respond with "0", "1", or "2". Do not include any additional text or explanations.} \\
        \bottomrule
    \end{tabular}
    \caption{LLM Prompts for \toxigen and \esnli.}
    \label{tab:LLM_prompts}
\end{table}

\clearpage

\end{document}